%% file: main.tex
\documentclass{article}

    \usepackage[preprint]{neurips_2020}

\usepackage[utf8]{inputenc} 
\usepackage[T1]{fontenc}    
\usepackage{hyperref}       
\usepackage{url}            
\usepackage{booktabs}       
\usepackage{amsfonts}       
\usepackage{nicefrac}       
\usepackage{microtype}      

\usepackage{times}

\usepackage{graphicx}
\usepackage{color}

\title{How to Close Sim-Real Gap? Transfer with Segmentation!}

\author{
Mengyuan Yan\\
School of Electrical Engineering\\
Stanford University\\
\texttt{mengyuan@stanford.edu}
\and
Qingyun Sun\\
School of Mathematics\\
Stanford University\\
\texttt{qysun@stanford.edu}
\and
Iuri Frosio\\
NVIDIA\\
\texttt{ifrosio@nvidia.com}
\and
Stephen Tyree\\
NVIDIA\\
\texttt{ styree@nvidia.com}
\and
Jan Kautz\\
NVIDIA\\
\texttt{jkautz@nvidia.com}
}
\date{}

\begin{document}
\maketitle

\begin{abstract}
One fundamental difficulty in robotic learning is the sim-real gap problem.
In simulation environment, we have infinite amount of data ( providing enough computing resource), additionally, we could generate good supervisions if needed for imitation learning, therefore, it is relatively easy to train a model-free control policy such as deep neural network parametrized control policy for a complex control tasks with unknown dynamic model; in real environment, we have extremely limited data as the acquisition of real data is slow and costly, additionally, we rarely have supervisions. Therefore, one natural attempt is to use control policy learned from simulation environment in real environment. However, there is always a gap on such transfer, especially when the deep control policy is end-to-end, namely, taking pixel as input. This gap is called sim-real gap. 

In this work, we propose to use segmentation as the interface between perception and control, as a domain-invariant state representation. We identify two sources of sim-real gap, one is dynamics sim-real gap, the other is visual sim-real gap. To close dynamics sim-real gap, we propose to use 
closed-loop control. For complex task with segmentation mask input, we further propose to learn a closed-loop model-free control policy with deep neural network using imitation learning. To close visual sim-real gap, we propose to learn a perception model in real environment using simulated target plus real background image, without using any real world supervision.

We demonstrate this methodology in eye-in-hand grasping task.
We train a closed-loop control policy model that taking the segmentation as input using simulation. We show that this control policy is able to transfer from simulation to real environment. The closed-loop control policy is not only robust with respect to discrepancies between the dynamic model of the simulated and real robot, but also is able to generalize to unseen scenarios where the target is moving and even learns to recover from failures. 
We train the perception segmentation model using training data generated by composing real background images with simulated images of the target. Combining the control policy learned from simulation with the perception model, we achieve an impressive $\bf{88\%}$ success rate in grasping a tiny sphere with a real robot.  
\end{abstract}



\input{Introduction}

\input{RelatedWork}

\input{Method}

\input{Results}

\input{Discussion}


\bibliographystyle{plainnat}
\bibliography{references}

\end{document}


\maketitle



\paragraph{Design of the imitation expert}
\label{subsec:expert}

We train our DNN controller in simulation, using Gazebo~\cite{Koe04} to simulate a Baxter robot.
The robot arm is controlled in position mode; we indicate with $[s_0, s_1, e_0, e_1, w_0, w_1, w_2]$ the seven joint angles of the robot arm and with $[g_0]$ the binary command to open or close the gripper.
Controller learning is supervised by an expert that, at each time step, observes joint angles and gripper state, as well as the position of the target sphere.
The expert implements a simple but effective finite-state machine policy to grasp the sphere.
In state $s_0$, the end-effector moves along a linear path to a point $6$cm above the sphere; in state $s_1$, the end-effector moves downward to the sphere; when the sphere center is within $0.1$cm from the gripper center, the expert enters state $s_2$ and closes the gripper. 
In case the gripper accidentally hits the sphere or fails to grasp it (this can happen because of simulation noise or control inaccuracies), the expert goes back to $s_0$ or $s_1$ depending on the new sphere position. 

Compared to a more elementary implementation of the expert directly aiming at the sphere, this expert is capable of grasping the sphere with a 96\%?? success rate (while the elementary expert achieves approximately 50\%). Failures are mostly associated with noise in simulation or inaccurate physics (we noticed that sometimes the sphere jumps away from the gripper abruptly when hit by it). This highlights the important of designing a proper expert for imitation learning - a policy which is ideal in theory may lead to sub-optimal results in simulation and subsequently invalidate the imitation learning procedure.

\paragraph{Image data generation}
To collect training data for the vision module, we need to have images as similar to real images as possible, while having cheap annotation of segmentation masks. We can obtain segmentation masks easily from Gazebo images. We can implement a color filter in HSV space that extracts all yellow pixels. However, as shown in insets of Fig. \ref{fig:teaser}, the Gazebo images are noise-free and well lit, shadows are poorly modeled especially when the end-effector is close to the sphere. They are significantly different from real images, thus directly training on these images will not generalize to real environments.

To make the images more realistic, we setup a scene in Povray~\cite{povray} with the same table, sphere, and camera arrangements as in Gazebo. Rendered images in Povray have much more realistic lighting, and segmentation masks can be directly rendered. We also approximate the gripper plates with two thin boxes to simulate shadows on the sphere. A sample of rendered image is shown in Fig.~\ref{fig:compose}.

Inspired by domain randomization~\cite{Jam17, Tob17}, to make the training images diverse, including various backgrounds and clutter objects, and at the same time more realistic, we alpha-blend rendered image patches of the sphere with random background images taken by the real robot camera. Fig.~\ref{fig:compose} demonstrates the composition process. We rendered $5000$ images of the sphere with random lighting condition and camera position, and collected $1600$ images of the real environment by setting the robot's arm at random poses in the robot's workspace. Random objects are placed in the robot workspace as clutter. The collection procedure is completely automated and takes less than an hour.
During training, the composed images are randomly shifted in the HSV space to account for the differences in lighting, camera exposure, and color balance. 

\paragraph{Close-loop controller learns to recover from failures}
In fact, imitation learning using only expert demonstrations may fail to capture such rare behavior, since the expert mostly succeeds at the first attempt and thus the training dataset would rarely include recovery behavior.
On the other hand, when training with DAGGER, at iteration $n$ the DNN controller executes the sub-optimal policy $\pi_n$ to collect new training data, so it can expose and learn to correct its own errors by querying the expert agent for advice.
As a result, the training data covers a larger state distribution than the distribution induced by expert demonstrations, and the trained agent effectively learns to recover from possible failures.

\paragraph{Comparisons with end-to-end approach}
\label{subsec:end2end}
Our two-module network can be trained separately or end-to-end. In the main paper we have reported the results while training the two modules separately. In this section we report the results training the network end-to-end, and compare with variants of our network similar to the ones used in ~\cite{Dev17,Sin17}.

We have made several modifications to the training process in order to make end-to-end training possible. First, at every time step, we copy the position of the end-effector camera and the sphere from Gazebo to Povray, and render an image to compose with a random background. While training the vision module independently, we can simply randomize the camera position to generate image-mask pairs, here we also need to make sure the produced image and segmentation mask matches the expert's joint commands. Second, since a rendering is required at every time step to replace the Gazebo image, we pause the physics simulation while rendering, in order the keep the command frequency the same as at test time. Last but not the least, we changed the number of training epoch from $200$ for every DAGGER iteration to $250 / i$ for the $i$th iteration, so that training time grows linearly with respect to the number of iterations, instead of quadratically. This reduced the end-to-end training time from estimated $134$ days to actually taking $7$ days on a single Titan X GPU.

In comparison, because the network is much smaller, training the controller module alone takes $12$ hours before modification, and can be reduced to $2$ hours after the modification without sacrificing performance. Training the vision module alone requires $1$ hour to collect the real background images, less than $1$ hour to render images with Povray paralleled over 4 threads, and $9$ hours to train the network. Because the vision module is supervised pixel-wise, each image is equivalent to $100 \times 100$ training samples, although strongly correlated. The network requires much less updates than the controller network to converge.

When testing on the real robot, The robot executes smoother trajectories, making less attempts before a successful one, and achieves $88\%$ success when no clutter objects present, following the evaluation process described in Sec. \ref{subsec:vision}. When clutter objects are present, the end-to-end trained network shows significant improvement in robustness, and accurately reaches the sphere in all $5$ trials, grasping the sphere in $4$ of them.

We also compare our method to two variants of our network. In the first variant, we keep the network structure unaltered, but do not give supervision to the segmentation layer, thus making it an unsupervised feature map. In the second variant, we increase the feature map from $1$ channel to $32$ channels and calculate the spatial softmax before concatenating with the robot joint angles and feeding to fully connected layers, similar to the networks used in ~\cite{Dev17,Sin17}. No supervision is given to the spatial softmax coordinates or the feature maps. Training is otherwise the same with our own network. 

\begin{figure}
\begin{center}
\includegraphics[width=0.7\linewidth]{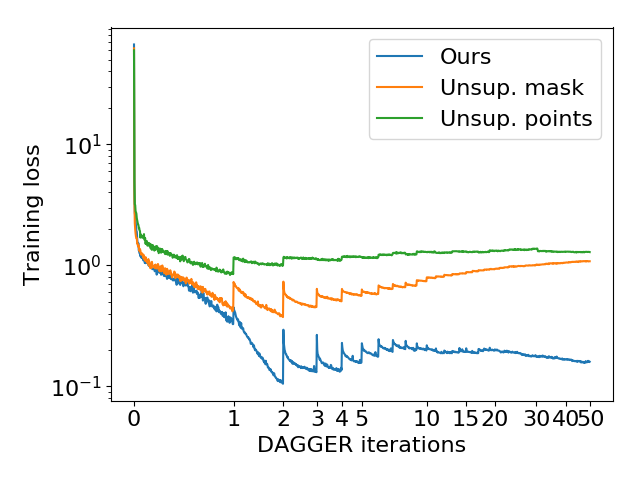}
\caption{Training loss for end-to-end training of several network variants. Ours (blue line) refers to our proposed network, with both expert action and segmentation supervision, and Povray + real background composed images as input. Unsupervised mask (orange line) refers to the same network and input images, but trained with only the expert action supervision, no segmentation supervision. Unsupervised points (green line) refers to network with spatial softmax on $32$ feature maps. The same input and expert action supervision is provided.}
\label{fig:end2endLoss}
\end{center}
\end{figure}

We plot the training losses in Fig~\ref{fig:end2endLoss}. Our modular network with additional segmentation supervision achieves significantly lower loss than the two variants. When testing on the real robot, our modular network can achieve $88\%$ success rate while the other two variants can drive the robot end-effector near the sphere, but are not accurate enough to grasp the sphere. Arguably this is not a completely fair comparison since addition supervision is used to train our modular network. However we want to demonstrate that by having a well-defined internal state representation as the interface between modules, it is possible to exploit the free annotations available in robotic simulators or graphics engines, and to train autonomous robots more efficiently and effectively. 

\bibliography{references}

%% file: Introduction.tex

\section{Introduction}
\label{sec:Introduction}
\paragraph{Segmentation as the interface between perception and control}
In robotic learning, it is common practice to resort to robotic simulators for the generation of training data, due to its safety and scalability.
However strategies learned in simulation have to generalize well to the real world, which justifies the research effort in the space of domain transfer~\cite{Tob17}. We take a slightly different approach in addressing the visual domain gap, which takes advantage of the asymmetric information between simulation and the real environment. Because all semantic and geometric information are freely available in simulation, a robot controller can be trained directly from this privileged information, faster and with a smaller neural network. Additional perception modules are trained to translate real sensor inputs into these representations, and they are not burdened with the need to understand the corresponding sensor measurements in simulation.

This work propose to decompose vision and control into separate network modules and use segmentation as their interface. Using segmentation as interface is beneficial in several ways. First, the segmentation mask eases human interpretation of robot behavior, and facilitates development and debugging. Second, perception and control modules can be trained separately or end-to-end. When training separately the vision module can use a variety of data sources collected offline, and from our experiments training separately is considerably faster. End-to-end fine-tuning can improve the performance further. Third, defining segmentation as the module interface effectively exploits the privileged information available in simulation as additional supervision. Finally, modular networks with a well-defined segmentation interface potentially allow the re-use of the same vision or controller module for different robots or environments.

We focuses on grasping with eye-in-hand camera, which is simpler in perception side because the occlusion problem is mostly avoided, but harder in control side, as a close-loop control is required. 

Training a closed-loop robot controller efficiently has its unique challenges. With closed-loop control, data collection depends on the policy network being trained, whether using reinforcement learning or imitation learning algorithms. If training with one particular network or hyper-parameters does not go well, the data collected cannot be easily reused to train the next improved version. Therefore, by choosing segmentation as the interface between perception and control, the perception modules can be trained as a prediction problem, where data is fully reusable and can be collected with massive parallel infrastructure; the controller module, taking this domain-agnostic mask as input and controlling robot joint angles, is trained efficiently in simulation using imitation learning and applied directly in real environments. Specifically, our vision module processes end-effector (eye-in-hand) camera images, and extracts the grasp target from the environment in the form of a segmentation mask. Training data for this module are generated by composing background images taken in the real environment with foreground objects from simulation.



Our real robot achieves $88\%$ success in grasping, only using RGB images from the end-effector camera and a closed-loop DNN controller.
The real robot is robust to clutter objects or a moving target, and has developed recovery strategies from failed grasp attempts.

\begin{figure}
\centering
\includegraphics[width=0.95\linewidth]{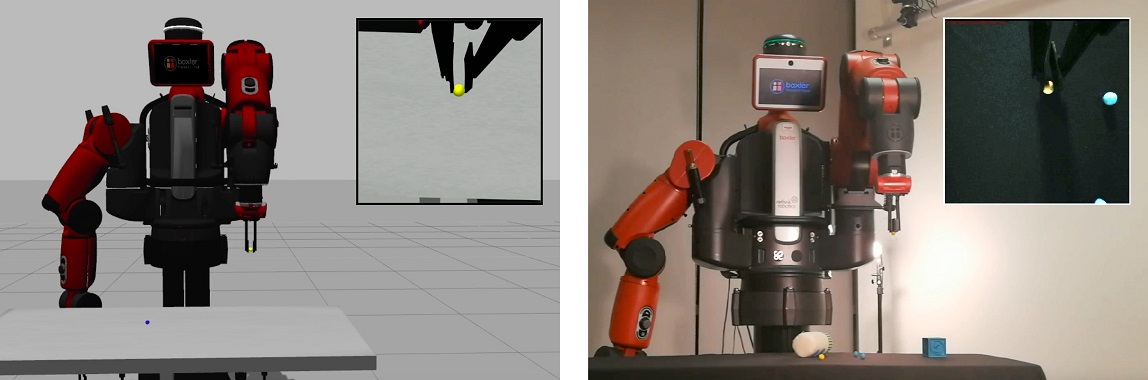}
\caption{We use imitation learning in simulation to train a closed-loop DNN controller, allowing a robot arm to successfully grasp a tiny (1.37cm of diameter) sphere (left panel).
The DNN controller's visual input is a binary segmentation mask of the target sphere, extracted from the RGB image captured by the end-effector camera (left inset).
A separate DNN vision module processes the real RGB images (right inset) to produce the same segmentation mask as in simulation, abstracting away the appearance differences between domains.
Combining the vision module and the controller module in a real environment (right panel), we achieve a grasping success rate of 88\% for the real robot.
}
\label{fig:teaser}
\end{figure}
\vspace{-10pt}

%
%
%


%% file: RelatedWork.tex

\paragraph{Related Work}
\label{sec:RelatedWork}
We review different approaches to learning-based robotic grasping system design. Some previous works~\cite{Mah16, Mah17, Jan17} use Convolutional Neural Networks (CNN) as grasp pose evaluators, and predict grasp success probability from visual observations and grasp poses. A separate grasp proposal generator is needed to work with the CNN evaluators. The CNNs observe one image and ranks the best grasp pose to be executed. Other works~\cite{Jam17} use end-to-end neural networks that directly map from visual input to robot joint commands, shifting the grasping system from open-loop to closed-loop. Although there are tools to visualize network filters and activations, it is often hard to decipher what the neural network has learned, or where the network is paying attention to for a specific decision. \citet{Dev17,Sin17} used network structures more similar to our work, where the network has a perception part and a controller part, with feature point coordinates as the connection between the two parts. The feature points are also internal state representations that humans can try to interpret, although their meaning is not always clear. In addition, because the internal state representations are only defined by format, not by content, the neural networks have to be trained end-to-end. In this paper, we also decompose the network into a perception part and a controller part, but we give specific meaning to the interface: it is defined as the segmentation of the target object. Combined with robotic simulation we are able to give direct supervision to the perception network and train the controller part independently from the perception part.



For the choice of the learning environment, robots can learn to grasp directly in the real world~\cite{Lev16, Sin17}, but training from scratch on real robots is costly, potentially unsafe, and requires very lengthy training time. 
Training in simulation is easier and cheaper, but transfer learning is needed to generalize from simulation to the real world.

Previous works had looked into the simulation-to-real transfer problem in grasping, for example~\cite{fang2018multi,fang2018learning}. The geometric configuration and appearance of the simulated environment can be extensively randomized, to create diverse training images that induce the neural network to have desired invariance to many visual aspects, and the neural network is shown to generalize also to unseen real environments~\cite{Jam17}.
~\citet{Ino17} used a variational autoencoder to change the style of the real images into that of the corresponding simulated images; although effective, this method requires coupled simulated and real image pairs to learn the style transfer mapping, thus it does not scale to complex environments.
In this paper, we compromise the reality of the real images in exchange for scalable data collection and free annotation of correspondences, and incorporate the idea of randomization to boost generalization. 
It is worth mentioning that the domain transfer problem does not apply only to vision: since the dynamics of a simulated robot may differ from its real-world counterpart, randomization can also be applied in this domain to facilitate generalization of the trained controller~\cite{Pen17, Tob17}.

Another design choice regards the learning algorithm.
The training speed depends on the cost of generating training data, the sample efficiency of the learning algorithm~\cite{Lev17}, and the balance of the available computational resources~\cite{Bab17}.
Deep RL algorithms have been successfully used to play Go and Atari games at superhuman level~\cite{Mnih15, Sil16, Sil17, Bab17}.
A3C is also employed for robotics in~\cite{Rus16}, although its low data efficiency is posing strong constraints.
More sample efficient RL algorithms, like DDPG~\cite{Lil15}, explore the solution space more effectively and consequently move some of the computational demand from the simulation to the training procedure, but still require a huge amount of data to reach convergence.
The most sample efficient learning procedures are instead based on imitation learning: in this case the trained agent receives supervision from human demonstrations or from an oracle policy, thus the need for policy exploration is minimal and sample efficiency is maximized~\cite{Jam17, Sin17}.
Many imitation learning variants have been proposed to improve test-time performance and prevent exploding error~\cite{Ros10, Ros11}. We used DAGGER~\cite{Ros11} to train our DNN controller in simulation, with an expert designed as a finite state machine.

Compared to the existing literature, our approach is different in a few ways. In grasping setting, we use the end-effector RGB camera, thus not requiring a fixed camera in a controlled pose; Our close-loop DNN controller allows correcting for pose estimation errors and dynamic changes of the environment in real time; our results also show that, although existing, the ``reality gap''~\cite{Pen17} between the dynamic model of the simulated and real robot is only a minor issue in the case of our close-loop controller.

%% file: Method.tex
\section{Method}
\label{sec:Method}
\paragraph{Deep closed-loop controller architecture and imitation learning}
\label{subsec:training}
\begin{figure*}
\begin{center}
\includegraphics[width=0.9\textwidth]{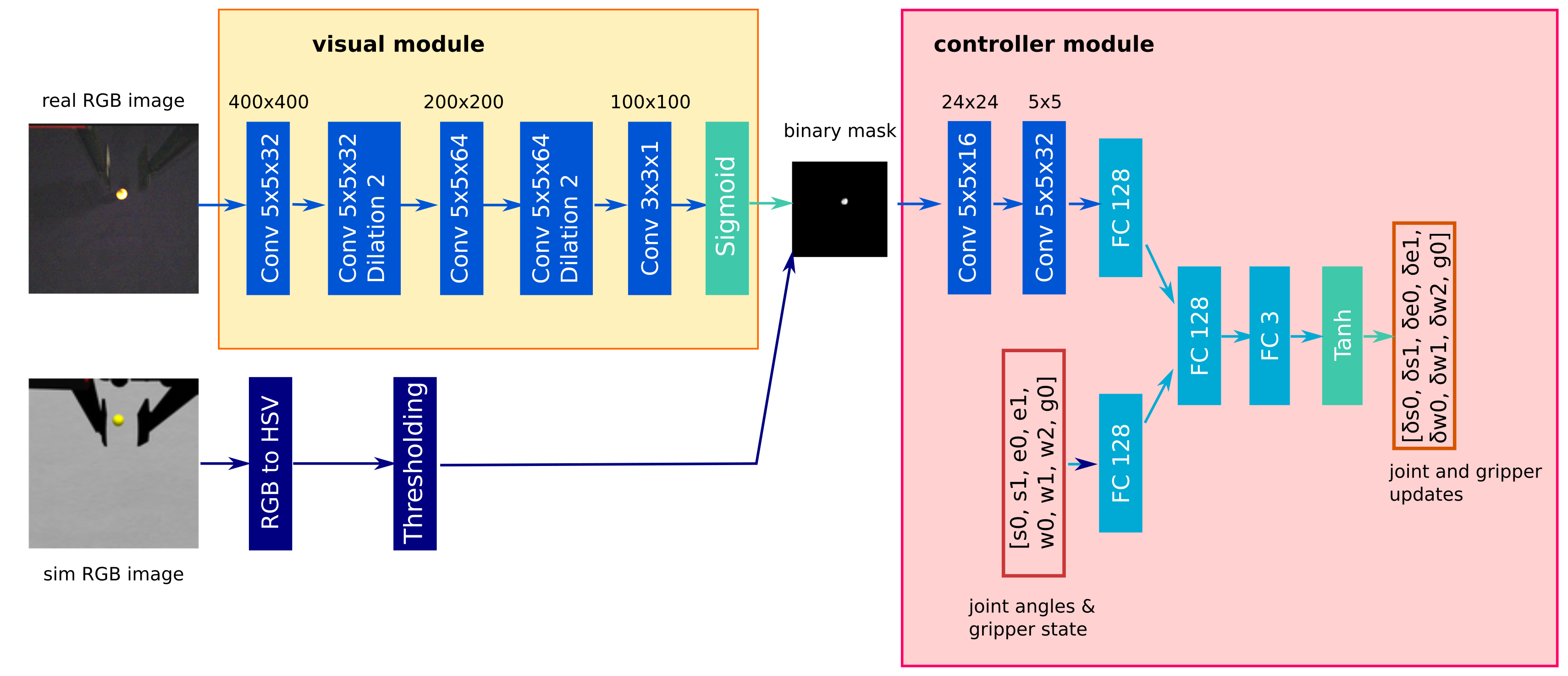}
\caption{The DNNs for the vision module (left) and closed-loop controller module (right). 
The vision module takes RGB images from the end-effector camera and labels the sphere pixels. The controller module  takes as input the segmentation mask and the current configuration of the robot arm (7 joint angles plus the gripper status), and it outputs an update for the robot configuration. 
}
\label{fig:network}
\end{center}
\end{figure*} 
\begin{figure}
\begin{minipage}{0.6\textwidth}
\includegraphics[width=0.9\linewidth]{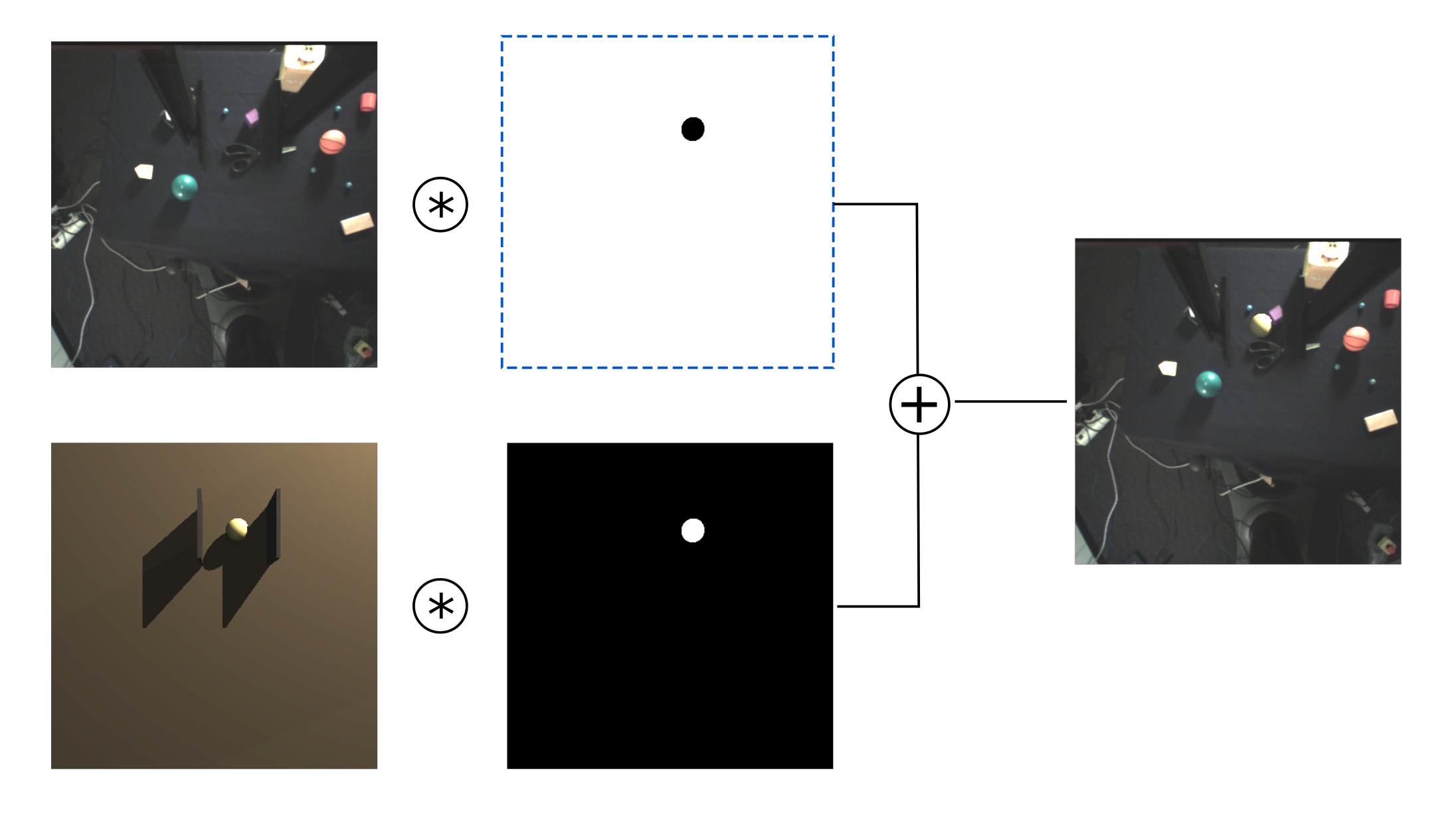}
\end{minipage}
\begin{minipage}{0.35\textwidth}
\caption{Composing Povray rendered sphere image (lower left) with real background image (upper left). The two images are blended according to the segmentation mask rendered from Povray (lower middle column), and the resulting image is on the right. All images are enhanced for viewing}
\label{fig:compose}
\end{minipage}
\end{figure}
Our closed-loop DNN controller processes the input information along two different pathways, one for the visual data and the other for the robot state.
The visual input is a $100 \times 100$ segmentation mask, obtained directly from simulation or from the output of the vision module, segmenting the target object from background.
The resulting field of view is approximately $80$ degrees.
The segmentation mask is processed by $2$ convolutional layers with $16$ and $32$ filters respectively, each of kernel size $5 \times 5$ and with stride $4$, and a fully connected layer with $128$ elements; ReLU activations are applied after each layer (see Fig. \ref{fig:network}).
The robot state pathway has one fully connected layer (followed by ReLU) that expands the $8$ dimensional vector of joint angles and gripper state into a $128$ dimensional vector.
The outputs of the two pathways are concatenated and fed to $2$ additional fully connected layers to output the action command, \emph{i.e.} the changes of joint angles and gripper status $[\delta s_0, \delta s_1, \delta e_0, \delta e_1, \delta w_0, \delta w_1, \delta w_2, g_0]$.
A \emph{tanh} activation function limits the absolute value of each command.
Contrary to~\cite{Jam17}, we do not use an LSTM module and do not observe a negative impact on the final result, although a memory module could help recovering when the sphere is occluded from the view of the end-effector camera.

We use DAGGER ~\cite{Ros11}, an iterative algorithm for imitation learning, to learn a deterministic policy that allows a robot to grasp a $1.37$cm diameter, yellow sphere.
Here we give a brief overview of DAGGER. We gave more details about how to generate the supervision for imitation learning from expert in the appendix in paragraph "Design of the imitation expert".

Given an environment $E$ with state space $s$ and transition model $T(\mathbf{s}, \mathbf{a}) \rightarrow \mathbf{s}'$, we want to find a policy $\mathbf{a} = \pi(\mathbf{s})$ that reacts to every observed state $\mathbf{s}$ in the same manner as an expert policy $\pi_E$. 

During the first iteration of DAGGER, we gather a dataset of state-action pairs by executing the expert policy $\pi_E$ and use supervised learning to train a policy $\pi_1$ to reproduce the expert actions.
At iteration $n$, the learned policy $\pi_{n-1}$ is used to interact with the environment and gather observations, while the expert is queried for the optimal actions on the states observed and new state-action pairs are added to the dataset.
The policy $\pi_n$ is initialized from $\pi_{n-1}$ and trained to predict expert actions on the entire dataset. \citet{Ros11} also discussed blending the learned policy $\pi_{n-1}$ with the expert policy when interacting with the environment, i.e. at each state, the action is calculated as $\beta_{n}\pi_E + (1-\beta_{n})\pi_{n-1}$. We choose $\beta_{n} = \delta(n=1)$, i.e. not blending expert policy with learned policies, since we do not see advantage of using $\beta_{n} = p^{n-1}, 0<p<1$. 
At each iteration $n>1$ the newly gathered state distribution is induced by the evaluated policy $\pi_{n-1}$, and over time the training data distribution will converge to the induced state distribution of the final trained policy. 

To train this DNN with DAGGER, we collect $1000$ frames at each iteration, roughly corresponding to $10$ grasping attempts.
At each iteration we run $200$ epochs of ADAM on the accumulated dataset, with a learning rate of $0.001$ and batch size $64$.
The training loss in DAGGER is the squared L2 norm of the difference between the output of the DNN and the ground truth actions provided by the expert agent, thus defined at iteration $n$ as:
\begin{equation}
\mathcal{L} = \sum_{\mathbf{s}}||\pi_n(\mathbf{s}) - \pi_E(\mathbf{s})||^2
\label{eq:L}
\end{equation}
%

%

\paragraph{Perception training using simulated target plus real background}
\label{subsec:transfer}
The vision module takes real images from the robot end-effector camera, and predicts segmentation masks of the target object. It is a visual domain translator from real environments to the simulated environment. However, it does not need to imagine the nuance appearance style of the particular simulator, such as the lighting condition, object color and texture. It only needs to keep the essential geometric information in the form of segmentation. The segmentation mask predicted by the vision module has to match the one obtained from the simulator, if the underlying robot pose and sphere position matches.

As shown in Fig. \ref{fig:network}, the vision module has $5$ convolution layers interleaved with $2$ pooling layers. It takes $400 \times 400$ RGB images and produces $100 \times 100$ segmentation masks. We use kernel size $5 \times 5$ and dilated convolution on the second and fourth convolution layers to have a large enough receptive field for each pixel prediction.

To ensure that the geometric information captured by the simulated and real cameras are coherent with each other, we calibrate the internal parameters (focal length, principal points) of the end-effector camera on the real robot and apply the same parameters in the Gazebo simulation.
We do not apply any distortion correction as we assume that a good policy executed with a closed-loop control tends to see the sphere in the center of the image, where distortions are minimal.

To collect training data for the vision module, we need to have images as similar to real images as possible, while having cheap annotation of segmentation masks. We can obtain segmentation masks easily from simulated images. We blend the rendered image patches of the sphere with random background images taken by the real robot camera. Fig.~\ref{fig:compose} demonstrates the composition process, see appendix in paragraph "Image data generation" for the details of our image domain randomization method to generate simulated target plus real background.

\begin{figure}
\begin{minipage}{0.65\textwidth}
\includegraphics[width=0.9\textwidth]{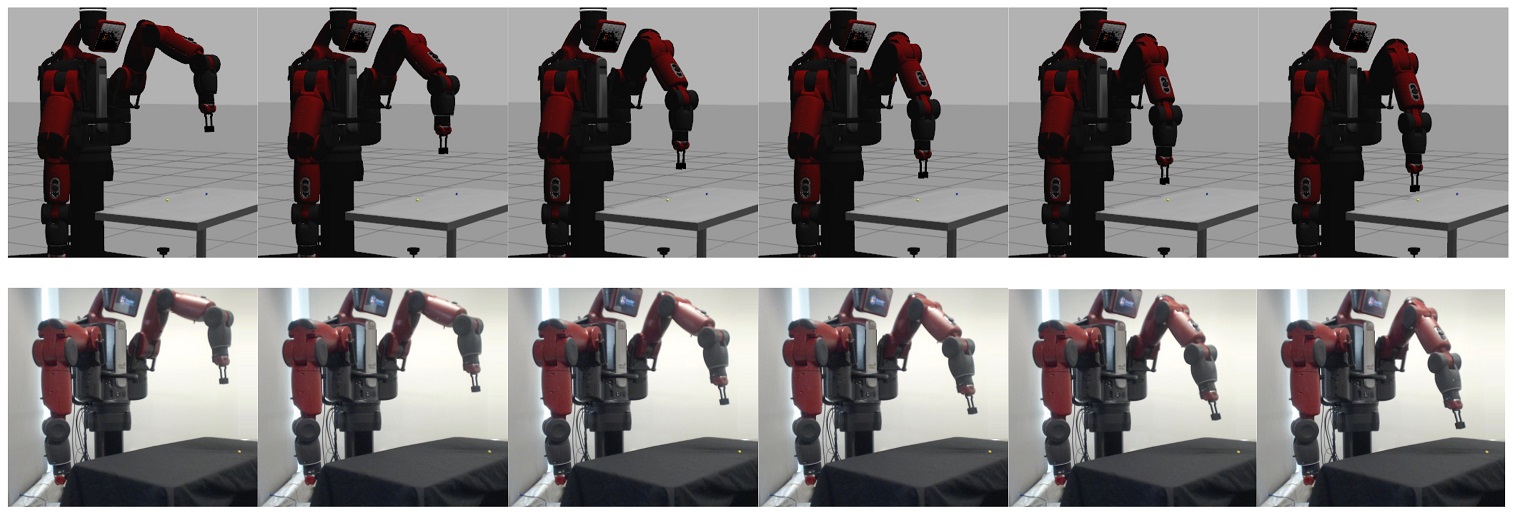}
\end{minipage}
\begin{minipage}{0.3\textwidth}
\caption{Starting from the same initial configuration (left-most panels), the simulated (top row) and real (bottom row) robots execute the same sequence of $150$ commands at $10$Hz. Because of differences in the dynamic model of the simulated and real robot, the final configurations of the two robots are different.}
\label{fig:control_issues}
\end{minipage}
\end{figure}

\paragraph{Dynamics domain transfer}
The ``reality gap'' between the dynamic response of the simulated and real robots may also require domain adaptation.
Several issues may contribute to the generation of this gap, including an inaccurate robot model or model parameter setting in simulation, hysteresis, joint frictions, delays in the transmission of the control signals, and noisy measurements of the state in the real robot~\cite{Tob17}. 
Fig. \ref{fig:control_issues} shows how the responses of the real and simulated robots can differ with an open-loop controller: starting from the same configuration, the execution of the same sequence of commands at the same frequency leads to two different robot configurations.
While small differences accumulate over time in the case of an open-loop controller, our choice of a closed-loop controller corrects execution errors online, leading to a stable and accurate system as shown in Section \ref{sec:Results}, without requiring any dynamic domain adaptation as in~\cite{Pen17}.

%% file: Results.tex

\section{Results and Discussion}
\label{sec:Results}
\begin{figure}
\begin{minipage}{0.6\textwidth}
\includegraphics[width=0.9\textwidth]{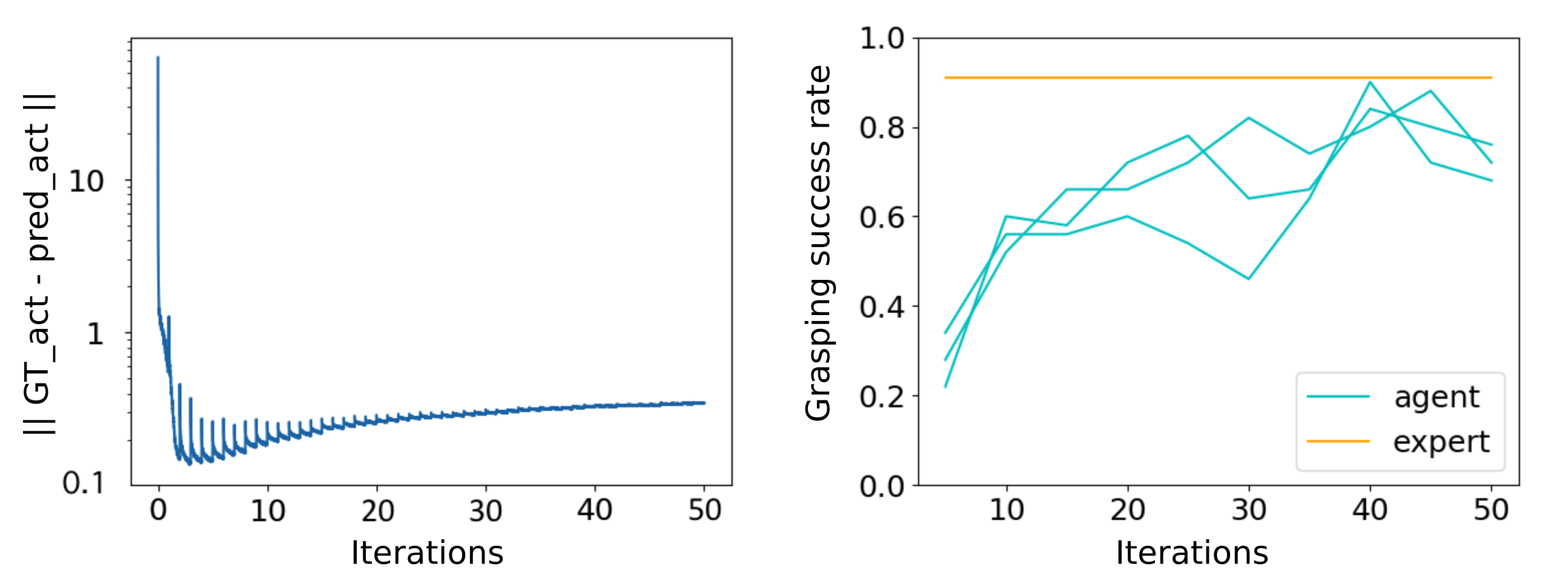}
\end{minipage}
\begin{minipage}{0.35\textwidth}
\caption{Left: DAGGER loss function $\mathcal{L}$ during training. Overfitting occurs in early iterations when the dataset is small; then the loss stabilizes around its optimal value. Right: grasping success rate of the DNN controller in simulation, evaluated three times every five iterations of DAGGER.}
\label{fig:simEval}
\end{minipage}
\end{figure}

\paragraph{Grasping in simulation}
\label{subsec:simulation}
We evaluate our DNN controller module in simulation at the end of each iteration of DAGGER.
Evaluation is performed by measuring the number of successful grasps for a sphere located at $50$ positions regularly spaced on a rectangular grid.
For each position, the trial ends with a successful grasp or after $150$ steps.
To account for uncertainties in the simulator, we run the evaluation three times.
Fig. \ref{fig:simEval} shows the grasping success rate as training progresses: $50$K training frames are sufficient to achieve a $92\%$ success rate, matching the performance of the expert.
Compared to~\cite{Rus16, Pop17, Jam17} that take $0.3$M, $50$M, and $1$M frames respectively to solve similar tasks, we see the superior data efficiency of DAGGER, relative to other reinforcement or imitation learning algorithms.

Visual inspection of failed attempts reveals that on rare occasions the grippers push against the table and cannot be closed; such cases could be solved if force sensors are available on the robot end-effector.
However, in most failure cases, the robot's end-effector reaches the sphere but touches it causing the sphere to roll away quickly out of the camera field of view, too far to be reached even for the expert. Such excessive speed of the sphere only appears in the simulator, probably due to the contrast in mass between the robot arm and the sphere, and a lack of good friction model. In the real environment, the sphere only moves slightly when the robot gripper touches it, and the robot has learned to recover by slightly raising its arm. 

\begin{figure}
\begin{minipage}{0.65\textwidth}
\includegraphics[width=0.9\textwidth]{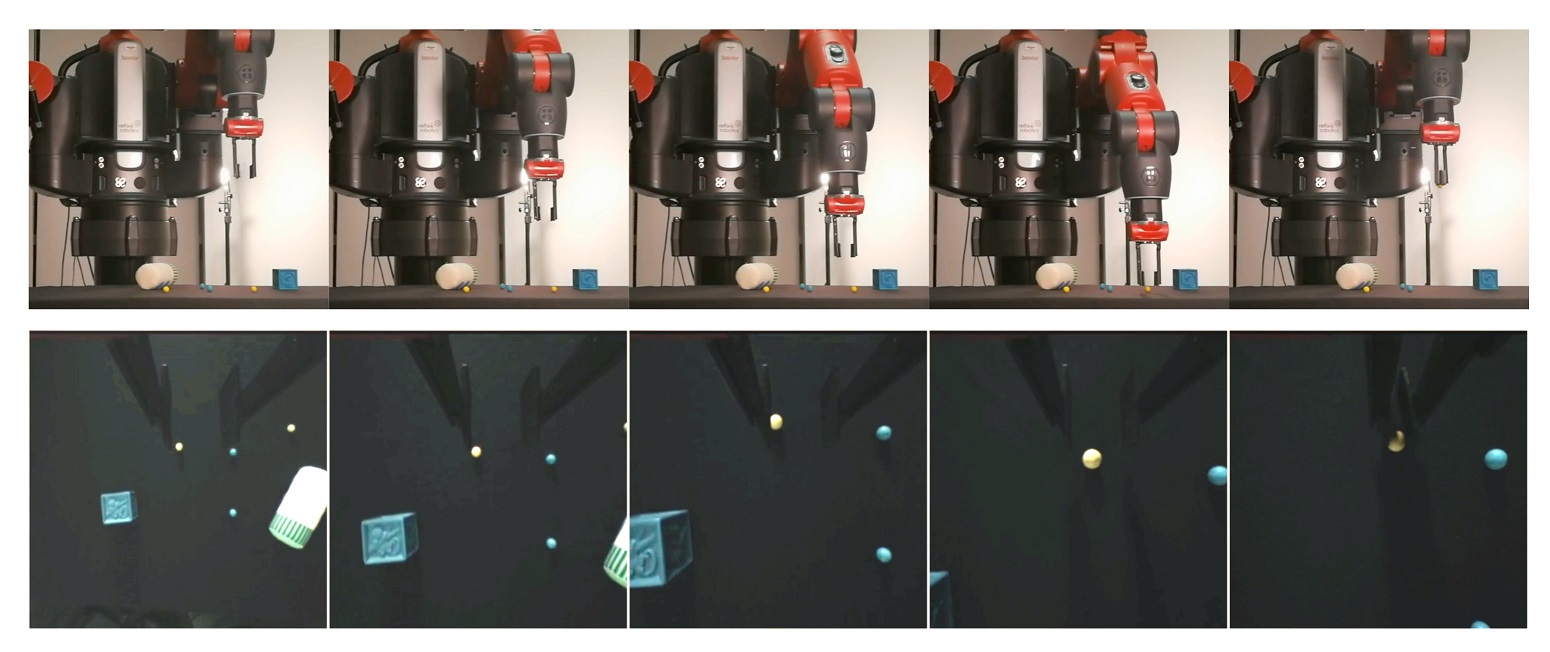}
\end{minipage}
\begin{minipage}{0.3\textwidth}
\caption{Snapshots of the learned agent grasping in a real environment. The only visual input of the DNN closed-loop controller is the end-effector camera image shown in the bottom row. We have modified the brightness and contrast of the images for ease of viewing.}
\label{fig:realResult}
\end{minipage}
\end{figure}

\paragraph{Transferring to a real robot}
\label{subsec:real}
There are two sim-real gaps, one is dynamics sim-real gap, another is visual sim-real gap.
The dynamics sim-real gap is naturally small, thanks to the closed-loop approach we take for the controller: since the controller corrects previous position errors, the ``reality gap'' between the dynamics of the simulated and real robots does not represent a critical issue, at least for a robot moving at limited speed. 

Another benefit of close-loop controller on the real robot is that we observe the emergence of recovery strategies from failed attempts, when the controller raises the end-effector slightly above the table to relocate the target sphere.
Such recovery behavior can only be scripted in the case of an open-loop formulation, as shown in~\cite{Lev16}, but it is learned automatically as an effect of the combined choices of a closed-loop controller, learning through DAGGER, and the design of the expert as a finite state machine. See paragraph "Close-loop controller learns to recover from failures" in appendix.


\label{subsec:vision}
\begin{figure}
\begin{minipage}{0.65\textwidth}
\includegraphics[width=0.9\textwidth]{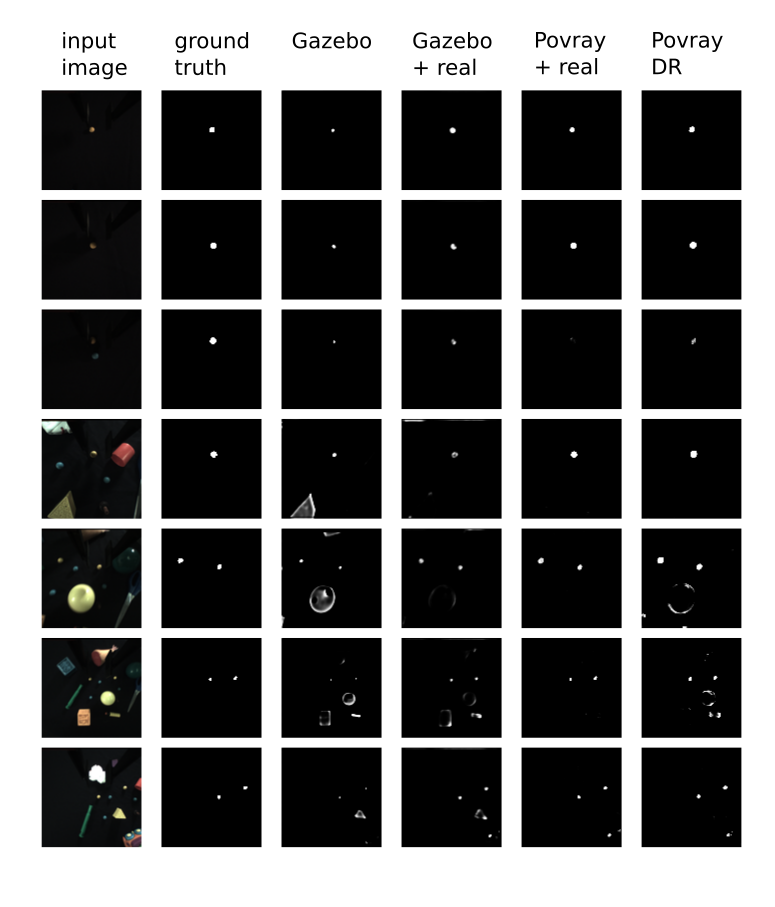}
\end{minipage}
\begin{minipage}{0.3\textwidth}
\caption{From left to right: input RGB images, ground truth segmentation masks, predicted segmentations from network trained with Gazebo images (at epoch 10), predicted segmentations from network trained with Gazebo + real images (at epoch 10), predicted segmentations from network trained with Povray + real images (our proposed method), and predicted segmentations from network trained with domain randomization implemented in Povray.}
\label{fig:segmentation_eval}
\end{minipage}
\end{figure}

\begin{figure}
\begin{center}
\includegraphics[width=0.45\linewidth]{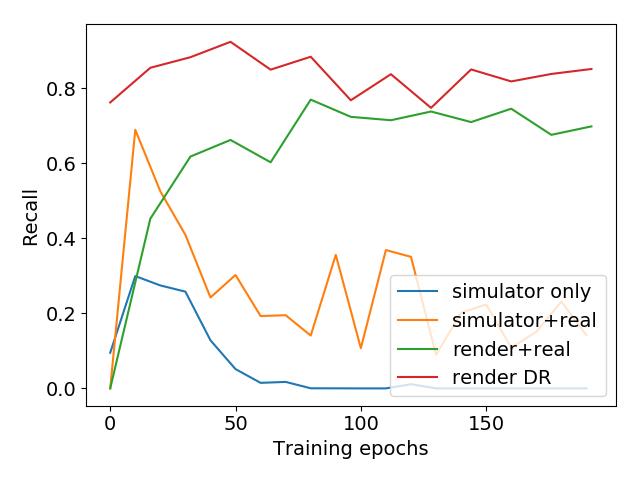}
\includegraphics[width=0.45\linewidth]{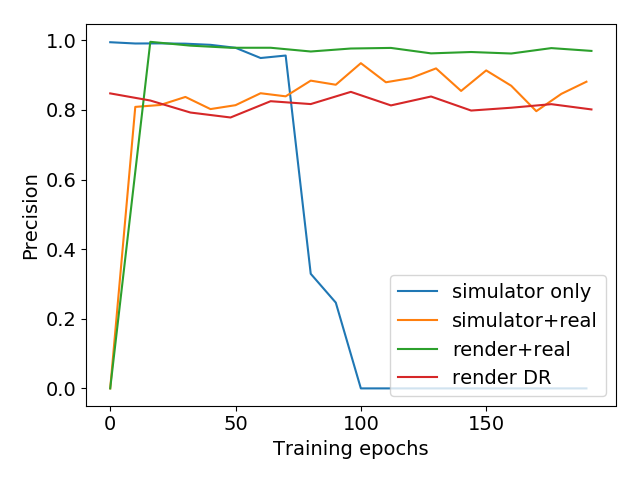}
\caption{Precision (right) and recall (left) of the vision module on the test set, when training on different datasets. Simulator only (blue line) refers to images collected from Gazebo. Simulator + real (orange line) refers to images composed of Gazebo images and real backgrounds. Render + real (green line) refers to images composed of Povray rendered spheres and real backgrounds. Render DR (red line) refers to Povray rendered images with domain randomization implemented.}
\label{fig:segmentationPR}
\end{center}
\end{figure}

An agent that is more robust to changes in the environment, \emph{e.g.} lighting conditions, is obtained with the introduction of the DNN vision module in Fig.~\ref{fig:network}. We compare the effectiveness of our proposed training dataset with other methods. As the first baseline we collected $5000$ images from Gazebo while the expert is in action, and used the images to train the same network. Second, we compose the above $5000$ images with the $1600$ real background images to train the same network. Third, we re-implemented domain randomization in Povray, generating geometric primitives with random shape, size, position, color, and solid or procedural texture. We also randomized lighting condition and camera viewpoints. We rendered $20000$ images to train the same network. For our own method, we rendered $5000$ images of the sphere in Povray, with randomized lighting condition and camera viewpoints, and composed them with the same $1600$ real background images.

We evaluate the DNN vision module on a set of $2140$ images from the real robot, collected by running the DNN controller in simulation and copying the sphere positions and robot trajectories to the real environment. Ground truth segmentation is annotated manually with the help of Maximally Stable Extremal Regions (MSER) detector~\cite{Mat04}. 

Precision and recall on the test set is evaluated every $10$ epochs during training and plotted in Fig.~\ref{fig:segmentationPR}. Because the lighting model in Gazebo renderer is rough, and lighting cannot be easily randomized, networks trained with Gazebo images, whether composed with real backgrounds, cannot generalize well to real images. Training the network using Povray rendered images achieves good results. Composing Povray rendered spheres and real backgrounds performs comparable to pure Povray images with domain randomization, with slightly lower recall but higher precision. 

Visual comparisons (Fig. \ref{fig:segmentation_eval}) confirms the clear advantage of using Povray rendered images compared to Gazebo images, due to more accurate lighting and the ability to depict shadows. The network trained by composing Povray images with real backgrounds occasionally fails to recognize the sphere from the image, while the network trained with domain randomization produces larger segmentation of the spheres, but also produces false positive predictions on the edges of other yellow objects. 

When the DNN controller uses the output of the DNN vision module, we evaluate the success rate of the real robot to grasp the sphere. We choose $5$ fixed positions to put the sphere, spanning the space used to training the DNN controller, and repeat grasping $5$ times at each position. Because the real robot moves slower than the simulated robot, a maximum of $250$ steps is allowed for each grasp. 

When no clutter objects are present, the real robot can achieve $84\%$ success rate while using the vision module trained with Povray-real composed images. Because the segmentation is not perfect especially when the gripper is close to the sphere, the robot will attempt to grasp several times before a final successful grasp. Using the vision module trained with domain randomization achieves a higher success rate of $88\%$.  

When clutter objects present near the sphere, a segmentation mask with false positives can drive the robot out of the desired trajectory, or even drive it out of the training distribution of the DNN controller, causing self-colliding movements.
With the vision module trained with domain randomization, the robot fails to grasp every time a yellow or orange object is present near the sphere. Since the network trained with Povray-real composed images suppress clutter objects much better, the robot succeeds in $2$ out of $5$ trials.

What if we training the policy end-to-end taking the image pixel space input? In the appendix of paragraph "Comparisons with end-to-end approach", we show that this method is significantly better than end-to-end approaches.

Snapshots from one successful grasp are shown in Fig.~\ref{fig:realResult}, while a video of the robot acting in the real environment can be seen in supplementary.

%% file: Discussion.tex

\section{Conclusion}
\label{sec:Discussion}
In this work, we propose a few principles to close the sim-real gap. 
\begin{itemize}
    \item First, we propose to use segmentation as the interface between perception and control. 
    \item Second, to close the visual sim-real gap, we propose to learn a perception segmentation model in real environment using simulated target plus real background image, without using any real world supervision.
    \item Third, to close the dynamics sim-real gap, we prose to use closed-loop controller. 
    \item Fourth, we show that imitation learning is a practical and model-agnostic way to learn a deep closed-loop model-free controller that takes segmentation mask as input.
\end{itemize}
 
The resulting system achieves a very significant $88\%$ success rate in grasping a tiny sphere on the real robot without any supervision or fine tuning from real environment. The system is robust to moving targets and background clutter and is often able to recover from failed grasp attempts.
